\documentclass[lettersize,journal]{IEEEtran}
\usepackage[nocompress,sort,space]{cite}
\usepackage{amsmath,graphicx}
\usepackage{makecell}
\usepackage{url}
\usepackage{amsmath}
\usepackage{amsfonts}
\usepackage{float}
\usepackage{hyperref}
\usepackage{url}
\usepackage{color}
\usepackage{algorithmic}
\usepackage{array}
\usepackage[table]{xcolor}
\usepackage{url}

\usepackage{pifont}
\definecolor{darkpastelgreen}{rgb}{0.01, 0.75, 0.24}
\definecolor{deeppink}{rgb}{1.0, 0.08, 0.58}
\newcommand{\cmark}{\textcolor{deeppink}{\ding{51}}}%
\newcommand{\xmark}{\ding{55}}%

\begin{document}

\title{StyleMorpheus: A Style-Based 3D-Aware Morphable Face Model}

\author{Peizhi~Yan,~\IEEEmembership{Student~Member,~IEEE,}
        Rabab~K.~Ward,~\IEEEmembership{Life~Fellow,~IEEE,}
        Dan~Wang,~\IEEEmembership{Student~Member,~IEEE,}
        Qiang~Tang,
        and~Shan~Du,~\IEEEmembership{Senior~Member,~IEEE}
\thanks{P. Yan, R. Ward, and D. Wang are with the Department
of Electrical and Computer Engineering, The University of British Columbia, Vancouver,
BC, Canada (Peizhi's e-mail: yanpz@ece.ubc.ca).}
\thanks{Q. Tang is with Huawei Technologies Canada, Burnaby, BC, Canada.}
\thanks{S. Du is with the Department of Computer Science, Mathematics, Physics and Statistics, The University of British Columbia (Okanagan), Kelowna,
BC, Canada.}
\thanks{Corresponding author: Shan Du (e-mail: shan.du@ubc.ca).}}


\maketitle

\begin{abstract}
For 3D face modeling, the recently developed 3D-aware neural rendering methods are able to render photorealistic face images with arbitrary viewing directions. The training of the parametric controllable 3D-aware face models, however, still relies on a large-scale dataset that is lab-collected. To address this issue, this paper introduces ``StyleMorpheus", the first style-based neural 3D Morphable Face Model (3DMM) that is trained on in-the-wild images. It inherits 3DMM's disentangled controllability (over face identity, expression, and appearance) but without the need for accurately reconstructed explicit 3D shapes. StyleMorpheus employs an auto-encoder structure. The encoder aims at learning a representative disentangled parametric code space and the decoder improves the disentanglement using shape and appearance-related style codes in the different sub-modules of the network. Furthermore, we fine-tune the decoder through style-based generative adversarial learning to achieve photorealistic 3D rendering quality. The proposed style-based design enables StyleMorpheus to achieve state-of-the-art 3D-aware face reconstruction results, while also allowing disentangled control of the reconstructed face. Our model achieves real-time rendering speed, allowing its use in virtual reality applications. We also demonstrate the capability of the proposed style-based design in face editing applications such as style mixing and color editing. Project homepage: \href{https://github.com/ubc-3d-vision-lab/StyleMorpheus}{\textcolor{deeppink}{\url{https://github.com/ubc-3d-vision-lab/StyleMorpheus}}}.
\end{abstract}

\begin{IEEEkeywords}
3D-aware face modeling, neural radiance fields, StyleGAN, parametric 3D face models.
\end{IEEEkeywords}

\section{Introduction}
\label{sec:intro}

\IEEEPARstart{T}{he} field of 3D-aware face modeling has witnessed rapid advancements in recent years, due to its promise in many applications such as virtual reality avatars \cite{bai2023high, ma2023otavatar}, face editing \cite{sun2022fenerf, jiang2022nerffaceediting}, and facial animation \cite{wu2022anifacegan, shen2023sd}. Different from traditional 3D face models which are mesh-based, 3D-aware methods do not explicitly represent the 3D face geometry (shape). This ability is enabled by employing neural radiance fields (NeRF) models \cite{mildenhall2020nerf, tewari2022advances, gao2022nerf} for learning an implicit 3D face representation. The key benefits of most 3D-aware face models over traditional 3D face models are (1) they are differentiable in nature and can learn from 2D images, (2) they can represent complex structures such as hair and eyeglasses, and (3) they can produce photorealistic appearance without the need for costly 3D modeling. Despite these benefits, the training of a 3D-aware parametric face model using in-the-wild face images (captured in uncontrolled, everyday environments, rather than in a structured studio setting) has remained a challenge. 

Amongst the traditional 3D face models, the most notable is the 3D Morphable Face Model (3DMM) \cite{blanz1999morphable} as it has the ability to independently and parametrically control the identity, expression, and appearance (color and lighting) of the reconstructed 3D face. In this work, we propose a 3D-aware face model that is parametrically controllable like 3DMM but only uses in-the-wild 2D face images for training. By leveraging a large amount of in-the-wild training data, our method can achieve better generalization ability and handle a wide range of diversity than using lab-collected data.

StyleGAN has achieved great success in 2D photorealistic face image synthesis using in-the-wild face images as training data \cite{karras2019style, karras2020analyzing}. It provides style-based control over generative image synthesis: StyleGAN leverages a high-dimensional latent space to manipulate the style aspects (such as age, gender, and other high-level facial attributes) of generated face images, emphasizing the separation of style from content to enable a more controllable image generation. To address the lack of 3D view consistency in 2D StyleGANs, 3D-aware style-based face modeling methods have recently been proposed. Some of these methods focus on random face image generation as in the original StyleGAN \cite{gu2021stylenerf, zhou2021cips, chan2022efficient, zhou2023cips}. Other methods focus on obtaining better controllability and view consistency in face image generation while allowing an explicit reconstructed 3D face geometry as the 3D prior \cite{sun2022controllable, xu2023omniavatar, sun2023next3d}. CGoF proposes the mesh-guided volume sampler to impose the shape information in volume rendering \cite{sun2022controllable}. OmniAvatar trains a Signed Distance Function (SDF) network to generate the head shape \cite{xu2023omniavatar}. This network is subsequently utilized to construct a volumetric correspondence map that serves as a guide for the process of volume rendering \cite{kajiya1984ray, max1995optical}. Next3D extends the idea of neural textures \cite{thies2019deferred} by synthesizing a neural texture through the StyleGAN generator \cite{sun2023next3d}. The synthesized neural texture and a reconstructed head mesh \cite{li2017flame} are then used in the rasterization process to derive a texture-rasterized field for 3D-aware rendering. Overall, these methods still have the problem of using explicit 3D face geometry. This dependency on an explicitly reconstructed 3D face shape in face reconstruction leads to high complexity in the model's architecture.

For face generation models to be parametrically controllable and to not employ explicit 3D face shapes, recent 3D-aware parametric face models \cite{zhuang2022mofanerf, hong2022headnerf, galanakis20233dmm} have drawn inspiration from the development of 3DMM \cite{blanz1999morphable}. MoFaNeRF is the first parametric 3D-aware face model trained on large-scale datasets of multi-view face images \cite{zhuang2022mofanerf}. Similar to the original NeRF, MoFaNeRF adopts volume rendering at the end to directly render the RGB image, which is computationally expensive and hinders its real-time application. HeadNeRF achieves real-time rendering by using NeRF to render low-resolution feature maps and having 2D neural rendering layers \cite{niemeyer2021giraffe, gu2021stylenerf} to generate high-resolution face images \cite{hong2022headnerf}. The parameter space of HeadNeRF is based on non-linear 3DMM \cite{tran2018nonlinear} and is adjusted through learning to capture more information such as hair, teeth, and detailed textures. Although both MoFaNeRF and HeadNeRF deliver impressive results, they heavily depend on large-scale lab-collected data, which is expensive to form and presents a challenge for generalizing to the in-the-wild images. 3DMM-RF explores the use of a 3DMM to synthesize images as training data and avoids the use of real images \cite{galanakis20233dmm}. However, 3DMM-RF does not fully take advantage of NeRF's benefits as it only mimics the 3DMM and thus cannot render hair and teeth. Importantly, due to the nature of NeRF which learns an implicit 3D representation, accurate camera pose is one of the key factors in training a NeRF to accurately render the details. When using in-the-wild images as training data, we can only roughly estimate the camera pose, which usually leads to blurry rendered images.

Recently, the 3D Gaussian Splatting (3DGS) \cite{kerbl20233d} based face avatars have shown success in rendering photo-realistic faces with controllability and real-time efficiency \cite{kabadayi2023gan, wang2023gaussianhead, qian2023gaussianavatars, xiang2023flashavatar, zhao2024psavatar, Yan_2025_WACV}. The 3DGS-based methods are somewhere between explicit and implicit 3D representations. However, these methods are not general 3D face models and focus on training a new face avatar model for each person, which is expensive. For example, GaussianAvatars needs to be trained for at least 6 hours for a single person to achieve satisfactory results \cite{qian2023gaussianavatars}. Moreover, the training data used in these methods are either multi-view videos \cite{kirschstein2023nersemble} or monocular videos \cite{zheng2022avatar} of the person with varying head poses and expressions. Our work has a different focus from the 3DGS-based face avatars as our method is a general 3D-aware face representation that can generate varying facial identities.

This work proposes an innovative style-based 3D-aware morphable face model, \textit{\textbf{StyleMorpheus}}, along with its training strategy. StyleMorpheus has four main advantages. First, it only uses in-the-wild 2D images for training. Second, it does not need an explicit 3D face shape as prior. Third, it is able to control face identity, expression, and appearance as in 3DMM. Fourth, it supports real-time 3D face rendering. StyleMorpheus also supports face reconstruction from a single-image input. A principal thrust of our effort is to overcome the reliance on large-scale synthetic or lab-derived data, a prevailing concern in existing 3D-aware parametric face models. To this end, we design a novel style-based auto-encoder. Our auto-encoder extracts the face-related style codes from the given face image. Following 3DMM, the style codes are separated into four semantic groups that can be controlled independently, namely identity (id.), expression (expr.), texture (tex.), and lighting (light). Our decoder is designed for real-time efficiency (anchored upon a low-resolution NeRF and 2D neural rendering). We use modulated layers as in StyleGAN to allow the decoder to apply the desired styles encoded in the style code to the feature maps, controlling the generated face.

Our training strategy has two stages. In the first stage, we train the auto-encoder to learn a representative latent style code space. In the second stage, we fine-tune the decoder with additional adversarial loss to improve the image quality. We conducted experiments to compare our model with existing 3D-aware parametric face models. Our model achieved state-of-the-art 3D-aware face image reconstruction performance.  Moreover, our network is lightweight and thus can run in real-time in inference time.

To summarize, our main contributions are:

\begin{itemize}

    \item We propose the StyleMorpheus 3D-aware face model, the first style-based neural 3DMM that can be successfully trained on in-the-wild 2D face images and supports various applications such as single-image 3D-aware face reconstruction and face editing.

    \item We propose a simple yet effective learning strategy that enables StyleMorpheus to learn a representative face style code space and to be trained on in-the-wild face images only.

    \item We propose a 3DMM-guided style code that allows StyleMorpheus to achieve outstanding disentangled (i.e., independent) control, demonstrated in the style mixing of faces. In addition, we design a novel facial part color editing application (e.g., changing the hair color) to further demonstrate the potential of our style-based face model.

\end{itemize}

The rest of this paper is organized as follows. In Section~\ref{sec:related}, we present a concise review of related works in the fields of 3D and 3D-aware face modeling. Section~\ref{sec:method} elaborates on the details of our proposed method. The experimental setups and results are presented in Section~\ref{sec:experiments}. Finally, Section~\ref{sec:conclusion} concludes this paper by summarizing our novel contributions and key findings.

\section{Related Works}
\label{sec:related}

\subsection{Parametric 3D Face Modeling}

Parametric 3D face modeling revolutionized the process of modeling a 3D face by using low-dimensional parameters to control the variations in the shape and appearance of the face \cite{vetter1998estimating, lewis2014practice, egger20203d}. The 3D Morphable Face Model (3DMM) is a type of parametric 3D face model that captures facial shape and texture variations across a population \cite{blanz1999morphable, paysan20093d, li2017flame}. 3DMMs are based on principal components learned from a large dataset of 3D facial scans. This allows the model to generate a wide range of facial expressions, appearances, and identities. 3DMM usually contains four groups of parameters: identity, expression, albedo, and lighting. The identity, expression, and albedo parameters are the blending weights of their corresponding principal components, while the lighting parameters are the coefficients of the spherical harmonics (SH) lighting, used to efficiently represent varying lighting effects \cite{sloan2023precomputed}. A popular downstream application of the 3DMM is single-image 3D face reconstruction. Its goal is to figure out the parameters that can faithfully recover the face's 3D shape and texture from the input image \cite{deng2019accurate, zhu2020reda, feng2021deca}. Reconstruction is also the prerequisite of other applications such as user-personalized face avatars \cite{bai2023high} and 3D face editing \cite{yan2022neo}.

\subsection{NeRF-Based 3DMM}

While 3DMM simplifies 3D face generation and enables easier control, it falls short of achieving photorealistic rendering capabilities. Also, 3DMM does not cover flexible and complex shapes such as hair and beard. On the contrary, NeRF does not need rigid 3D geometries but can synthesize 3D-consistent (i.e. 3D-aware) and photorealistic images. To achieve both controllability and realistic rendering, recent works propose to learn a 3DMM using NeRF \cite{zhuang2022mofanerf, hong2022headnerf, galanakis20233dmm}. MoFaNeRF is the first NeRF-based 3DMM \cite{zhuang2022mofanerf}. It relies on large-scale lab-collected facial images with multiple views and expressions for each subject as training data. The NeRF used in MoFaNeRF is conditioned on shape, expression, and texture parameters derived from the dataset. Similar to MoFaNeRF, HeadNeRF utilizes the large-scale lab-collected facial images as training data and uses the pre-fitted 3DMM coefficients as the initial face code to train a NeRF-based model. It further enhances the representation of the face code by fine-tuning both the face code and the model using in-the-wild face images \cite{hong2022headnerf}. 3DMM-RF learns on the synthetic images rendered from synthesized 3DMM faces and achieves superior generalization ability in fitting on in-the-wild face images \cite{galanakis20233dmm}. Whereas, 3DMM-RF inherited the problem of not modeling hair from a traditional 3DMM. In contrast to the above-mentioned methods, our approach exclusively uses in-the-wild images for training, enabling greater diversity and reduced cost. This approach also allows us to effectively capture diverse hairstyles, setting our method apart from MoFaNeRF and 3DMM-RF.

\begin{figure*}[ht]
    \centering
    \includegraphics[width=0.8\linewidth]{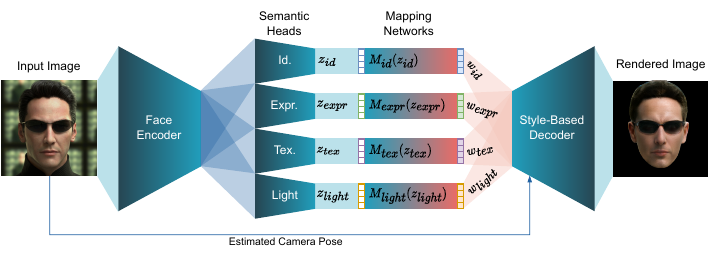}
    \caption{The proposed model architecture follows an auto-encoder structure. The encoder consists of an ArcFace-based face image encoder and four semantic heads to produce identity, expression, texture, and lighting code $z$s in $\mathcal{Z}$ space. Following the semantic heads are mapping networks $M$s that map the corresponding style code to $w$s in $\mathcal{W}$ space. The style codes in $\mathcal{W}$ space are then used in the proposed style-based decoder to reconstruct the input image.}
    \label{fig:framework}
\end{figure*}

\subsection{Style-Based 3D-Aware Face Models}

Generative adversarial networks (GANs) \cite{goodfellow2014generative} have played an important role in face image generation \cite{karras2019style, liu2022isf, hang2023language, zhou2019branchgan}. Based on GANs, StyleGAN introduces several innovative techniques to stabilize the training process and improve the controllability of face image generation \cite{karras2019style, karras2020analyzing, karras2021alias}. These include progressive training that allows photo-realism in high-resolution \cite{karras2017progressive}, the mapping network that enables a less entangled intermediate latent space \cite{karras2019style}, and style-based modulated convolution \cite{karras2020analyzing} that adjusts the statistics of intermediate feature maps to control the ``style'' \cite{huang2017arbitrary} of generated image. Building upon these advancements, recent works have extended the frontier of style-based 2D generative face models into the realm of 3D-aware synthesis: 

Pi-GAN \cite{chan2021pi} proposes a modulated version of the SIREN layer \cite{sitzmann2020implicit} to replace the MLP layers used in the original NeRF, resulting in a style-based NeRF network. This allows the model to generate view-consistent face images conditioned on both the style code and camera pose. Methods such as StyleNeRF \cite{gu2021stylenerf} and CIPS-3D \cite{zhou2021cips, zhou2023cips} only use the style-based NeRF to render a low-resolution feature map, then utilize upsampling and 2D style-based layers to gradually increase the resolution. This coarse-to-fine strategy significantly reduces the computational cost. EG3D \cite{chan2022efficient} presents tri-plane NeRF, which forms a tri-plane radiance field representation by reshaping the 2D feature maps generated by StyleGAN. Like the original StyleGAN, these methods randomly sample style codes to synthesize the face images, making it hard to control the generation. In our work, the style codes are disentangled through learning, allowing our generator to gain more controllability.

In recent studies, researchers have introduced 3D prior conditioned face models that combine both 3D-aware and traditional 3D representations \cite{sun2022controllable, xu2023omniavatar, sun2023next3d}. Utilizing a 3D prior offers the advantage of capturing shape-related deformations like facial expressions through explicit 3D geometries, which simplifies expression control. However, these approaches aren't exclusively implicit 3D rendering techniques since they still rely on explicit 3D face/head geometries during rendering, implying a reliance on accurate 3D face reconstructions. Furthermore, combining 3D geometry with 3D-aware rendering increases model complexity and impedes real-time efficiency. In contrast, our method does not use explicit 3D shapes, allowing a more efficient face rendering and simpler face reconstruction process.

\section{Method}
\label{sec:method}

\subsection{Preliminary: 3D Morphable Face Model}

The 3D morphable face model (3DMM) \cite{blanz1999morphable} uses 3D textured mesh for the 3D face representation. Since the scanned 3D faces are registered to a common mesh topology, the shape of a registered 3D face can be represented as a matrix $ S \in \mathbb{R}^{V \times 3}$ of 3D mesh vertex locations, where $V$ is the total number of vertices. A face shape can be synthesized through the following shape model formula \cite{li2017flame}:
\begin{equation}
    S_{model} = S_{mean} + c_{id} \mathbf{S}_{id} + c_{expr} \mathbf{S}_{expr},
    \label{eq:3dmm_shape_model}
\end{equation}
where $S_{mean}$ is the average face shape across the dataset, $\mathbf{S}_{id} \in \mathbb{R}^{P_{id} \times V \times 3}$ and $\mathbf{S}_{expr} \in \mathbb{R}^{P_{expr} \times V \times 3}$ are the first $P_{id}$ and $P_{expr}$ principal components of identity and expression residuals derived from the dataset, $c_{id} \in \mathbb{R}^{P_{id}}$ and $c_{expr} \in \mathbb{R}^{P_{expr}}$ represent the corresponding coefficients that control the synthesized face shape.

3DMM uses vertex color values to represent the texture of a 3D face, without considering any shading, shadows, or other lighting effects \cite{paysan20093d}. Similar to the shape model, the texture model is formulated as:
\begin{equation}
    T_{model} = T_{mean} + c_{tex} \mathbf{T},
    \label{eq:3dmm_color_model}
\end{equation}
where $T_{mean} \in \mathbb{R}^{V \times 3}$ are the average vertex color values across the dataset, $\mathbf{T} \in \mathbb{R}^{P_{tex} \times N \times 3} $ are the first $P_{tex}$ principal components of the vertex color residuals, and $c_{tex} \in \mathbb{R}^{P_{tex}}$ represent the texture coefficients. The lighting effects are approximated by the SH lighting \cite{sloan2023precomputed}, with coefficients $c_{light}$ \cite{deng2019accurate, feng2021deca}.

\subsection{Model Architecture}

The representation ability of the original 3DMM codes is limited mainly by two reasons. First, the 3DMM does not model the hair and teeth due to the complexity of modeling them in 3D. Second, the vertex color-based texture model is low-resolution, which cannot accurately represent complex textures such as facial hair. Therefore, we design an auto-encoder architecture (see Fig.~\ref{fig:framework}) to enable learning a more representative code space. We use the pre-trained ArcFace \cite{deng2019arcface} as the backbone of our face encoder. We connect four MLP-based networks to the face encoder, as semantic heads to derive the disentangled style codes in space $\mathcal{Z}$: $z_{id} \in \mathbb{R}^{100}, z_{expr} \in \mathbb{R}^{79}, z_{tex} \in \mathbb{R}^{100}, z_{light} \in \mathbb{R}^{27}$. The dimensions of our style codes $z$s are aligned with non-linear 3DMM \cite{tran2018nonlinear}. Different from the StyleGAN's $\mathcal{Z}$ space which is a unit normal distribution $\mathcal{N}(0,1)$, our $\mathcal{Z}$ space is learned to be disentangled. Following StyleGAN, we use mapping networks to map the style code from $\mathcal{Z}$ space to $\mathcal{W}$ space. The dimensions of $w$s are the same as $z$s.

\begin{figure*}[ht]
    \centering
    \includegraphics[width=\linewidth]{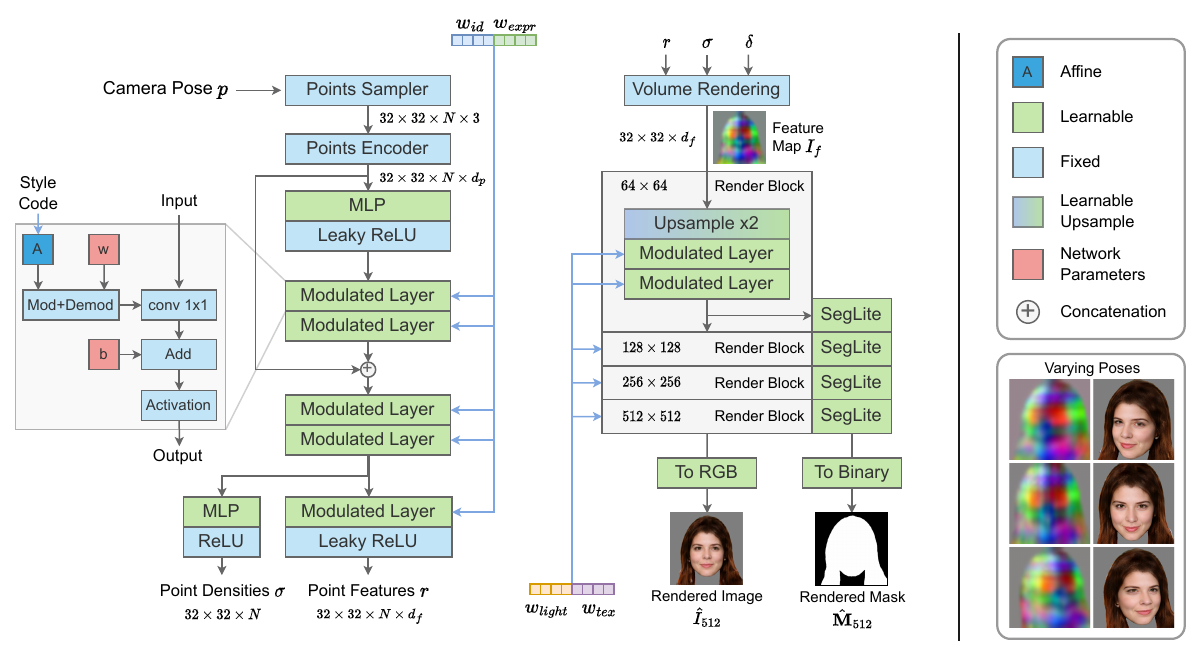}
    \caption{The proposed style-based decoder architecture consists of a low-resolution NeRF to produce the 3D-aware feature map (left) and a set of render blocks (right) to produce the high-resolution RGB face image. For convenience, we only show the last ``To RGB" layer in this figure. We use modulated layers in both NeRF and render blocks to accept style codes. The identity and expression codes are concatenated to give to the NeRF's modulated layers. The texture and lighting codes are concatenated and used in the render blocks to control the appearance of the generated face.}
    \label{fig:decoder}
\end{figure*}

Our style-based decoder follows a coarse-to-fine strategy to achieve real-time rendering \cite{gu2021stylenerf, hong2022headnerf}: as NeRF suffers from high computational cost in rendering high-resolution images, we only use a NeRF-based renderer to render the low-resolution feature map $I_{f} \in \mathbb{R}^{32 \times 32 \times d_f}$ at $32 \times 32$ resolution, where $d_f$ is the number of feature channels. We use $d_f = 256$ in our experiment. Then we apply four render blocks to generate the RGB image $\hat{I}_{512} \in \mathbb{R}^{512 \times 512 \times 3}$ at $512 \times 512$ resolution. We implement the modulated layer by using the weight-demodulated 2D convolution layer \cite{karras2020analyzing} with $1 \times 1$ convolution filters. It is demonstrated that $1 \times 1$ convolution layer is rotation-and-shift equivariant which is important in the 3D-aware setting \cite{karras2021alias, gu2021stylenerf}. We use the modulated layer in both NeRF and render blocks to allow them to accept style codes.

We formulate the camera pose as $p \in \mathbb{R}^{6}$, where the first three values are yaw, pitch, and roll angles for the 3D rotation $\mathbf{R} \in SO(3)$ and the last three values represent the 3D translation $\mathbf{t} \in \mathbb{R}^{3}$. Our camera model uses perspective projection. Given a camera pose $p$ we march $32^2$ rays through a $32 \times 32$ pixel grids and sample $N$ points along each ray. Positional encoding is used to map the 3-dimensional point locations to $d_p > 3$ dimensions to capture high-frequency content \cite{mildenhall2020nerf}. When applying our modulated layer on the point features, we reshape the point features to $32^2 \times N \times d_p$ so that the $1 \times 1$ 2D convolution acts like a fully connected operation ``channel-wise". The modulated layers in our NeRF use the sine activation function following SIREN \cite{sitzmann2020implicit}, to improve the ability to model high-frequency information. Since the NeRF is low-resolution, we only expect it to learn shape-related representation. For this reason, we only apply $w_{id}$ and $w_{expr}$ to our NeRF's modulated layers. The outputs of our NeRF are the estimated densities $\sigma \in \mathbb{R}^{32 \times 32 \times N}$ and features $r \in \mathbb{R}^{32 \times 32 \times N \times d_f}$ of the sampled points. The feature map $I_f$ is rendered through volume rendering \cite{kajiya1984ray, max1995optical}. For clarity, we reformulate the volume rendering equation in the following discrete format:
\begin{equation}
    I_f(x,y,k) = \sum_{i = 1}^{N-1}  t(x,y,i) \alpha(x,y,i) r(x,y,i,k),
    \label{eq:volume_rendering}
\end{equation}
where $t(x,y,i)$ is the accumulated transmittance from the first point to the $i^{th}$ point along the ray originating from the camera and passes through the pixel location at $(x,y)$, $\alpha(x,y,i)$ is the corresponding alpha compositing value. Theoretically, one needs to ensure $N>1$ when sampling the points. But since Equation~\ref{eq:volume_rendering} is an approximation of the continuous volume rendering function, the larger $N$ leads to a more accurate approximation. Equations \ref{eq:transmittance} and \ref{eq:alpha_composite} are the computations of $t(x,y,i)$ and $\alpha(x,y,i)$, where $\delta(x,y,i)$ represents the distance between the $i^{th}$ and $(i+1)^{th}$ adjacent points on the ray passes through $(x,y)$.

\begin{equation}
    t(x,y,i) = 
    \begin{cases}
    exp[- \sum_{j = 1}^{i-1} \sigma(x,y,j) \delta(x,y,j) ] & \text{if } i > 1,\\
    1.0 & \text{if } i = 1.
    \end{cases}
    \label{eq:transmittance}
\end{equation}

\begin{equation}
    \alpha(x,y,i) = 1 - exp[- \sigma(x,y,i) \delta(x,y,i) ].
    \label{eq:alpha_composite}
\end{equation}

We use four render blocks to increase the resolution and control the texture and lighting effects of the final rendered image. In each render block, we start with a $2\times$ learnable upsampling layer based on PixelShuffle \cite{shi2016real} and a fixed blur kernel \cite{zhang2019making}. Followed by the upsampled feature map are two modulated layers, and we use LeakyReLU as the activation function. Each render block has an attached $1 \times 1$ 2D convolution layer to convert the layer's output to an RGB image $\hat{I}_{res}$ ($res$ is the corresponding block's resolution). We apply light-weighted SegLite blocks \cite{yan2023seglite} that utilize the feature maps from different rendered resolutions, to generate the binary head mask $\hat{\mathbf{M}}_{512} \in \mathbb{R}^{512 \times 512}$. Please refer to Fig.~\ref{fig:decoder} for our detailed decoder architecture.

\subsection{Loss Function}

The proposed framework mainly learns to reconstruct the input face image. Following deep learning single-image face reconstruction methods \cite{deng2019accurate, feng2021deca, lin2022high, tu20203d}, we use photometric loss and perceptual loss as our reconstruction loss. The photometric loss is also known as pixel loss, which is the averaged pixel-level distances between the reconstructed and target images. Inspired by progressive training \cite{karras2017progressive}, we compute L1 distance at different resolution scales from $64 \times 64$ up to $512 \times 512$ to improve the convergence speed and encourage each render block to learn sufficient information. Denote $I$ as the ground truth image with resolution greater than or equal to $512 \times 512$, and its corresponding head mask as $\mathbf{M}$ (including face, hair, and ears), our multi-resolution photometric loss is defined as:
\begin{equation}
    \mathcal{L}_{photo} = \sum_{i = log_{2} 64}^{log_2 512}  \gamma_{2^{i}} \| \hat{I}_{2^{i}} - \mathbf{M}_{2^i} \odot I_{2^{i}} \|_{1} ,
    \label{eq:photo_loss}
\end{equation}
where $\gamma$s are the weighting factors, $\odot$ is the Hadamard product. We use bi-cubic interpolation to down-scale $I$ and $\mathbf{M}$ to match the resolution of each render block. 

The perceptual loss captures the semantic content of images, encouraging generated images to have similar high-level features as target images. We use LPIPS loss \cite{zhang2018unreasonable} as our perceptual loss $\mathcal{L}_{perc}$ between $\hat{I}_{512}$ and $\mathbf{M}_{512} \odot I_{512}$. The LPIPS loss implementation we use is from \footnote{\url{https://github.com/richzhang/PerceptualSimilarity}}, and we use its pre-trained VGG-16 network \cite{simonyan2014very} as the feature extractor. We also tested the identity loss \cite{deng2019accurate, galanakis20233dmm} in our experiment. Its effect is limited in our setting, thus we do not use the identity loss in our reconstruction loss. Our reconstruction loss is defined as:
\begin{equation}
    \mathcal{L}_{recon} = \lambda_{photo}\mathcal{L}_{photo} + \lambda_{perc}\mathcal{L}_{perc}, 
    \label{eq:recon_loss}
\end{equation}
where $\lambda$s are loss term weights.

The reconstruction loss only encourages the decoder to generate the face image with a black background. We use $L_{seg}$ which is the averaged pixel-level cross-entropy loss between $\hat{\mathbf{M}}_{512}$ and $\mathbf{M}_{512}$ to train the light-weighted SegLite-based segmentation branch. The SegLite-based segmentation branch is useful in removing the rendered background without affecting real-time efficiency.

Following \cite{hong2022headnerf}, we compute the code distance between $z$s and the estimated non-linear 3DMM \cite{tran2018nonlinear} coefficients $c$s, to ensure the code space $\mathcal{Z}$ is disentangled and aligned with 3DMM. This code loss term is used as a regularization:
\begin{equation}
    \mathcal{L}_{code} = \sum_{g} \lambda_{g} \| z_{g} - c_{g} \|_{2}^{2} ,
    \label{eq:code_regulation}
\end{equation}
where $g \in \{id, expr, tex, light\}$, and $\lambda$s are the corresponding loss term weights. 

Because in-the-wild images have been used as training data, we can only estimate the camera pose for each image. The estimation is done by fitting a 3DMM face to the image and using the optimized camera pose. In this case, we have to assume the camera intrinsics are the same across all the images. However, in reality, in-the-wild images are usually taken by cameras with different optical configurations. Moreover, fitting a 3DMM to a single image also encounters the unavoidable problem of inaccurate reconstruction. Therefore, the camera poses are inaccurate for in-the-wild images, which causes the learned 3D-aware model to generate blurry images. To address this problem, we use adversarial learning to improve our style-based decoder's generation quality and to encourage the decoder to learn to generate high-frequency details. Denote $E$ as our face image encoder, $G$ as the generator (mapping networks and style-based decoder), and $D$ as the discriminator network. In terms of adversarial learning \cite{goodfellow2014generative}, our objective is to solve the following minimax game:
\begin{equation}
    \underset{G}{\text{min}} ~ \underset{D}{\text{max}}  \underset{(I,p) \sim \text{data}}{\mathbb{E}} \{  logD(I) + log[1 - D(G(E(I), p))] \} ,
    \label{eq:minimax}
\end{equation}
where $p$ is the estimated camera pose of $I$.

Based on Equation~\ref{eq:minimax}, the discriminator loss is defined as:
\begin{equation}
    \begin{aligned}
    \mathcal{L}_{D} &= log ( 1 + exp( D(G(E(I), p)) ) )  \\
    &+ log( 1 + exp(-D(I)) ) + \mathcal{L}_{R1},
    \end{aligned}
    \label{eq:discriminator}
\end{equation}
where $\mathcal{L}_{R1}$ is the R1 regularization loss \cite{mescheder2018training}. The generator loss is defined as:
\begin{equation}
    \mathcal{L}_{G} = log(1 + exp( - D( G(E(I),p) ) ) ).
    \label{eq:generator}
\end{equation}

\subsection{Training Strategy}

Our training strategy has two stages. In the first stage, we train the auto-encoder end-to-end. Because the face encoder is a pre-trained ArcFace network, in the first training epoch, we freeze the face encoder (not including the semantic heads) and train the rest of the network. Starting from the second training epoch, we train the entire network. The objective of this stage is: 
\begin{equation}
    \underset{\theta_{E},\theta_{G}}{argmin} ~ \mathcal{L}_{recon} + \mathcal{L}_{code} + \lambda_{seg}\mathcal{L}_{seg},
    \label{eq:ae_loss}
\end{equation}
where $\theta$ represents the network parameters. Note that, we don't use the segmentation loss term $\mathcal{L}_{seg}$ until the training converges to a stable state.

In the second stage, we fine-tune the generator with adversarial loss. We train the generator and discriminator alternately in the same way as StyleGAN2 \cite{karras2020analyzing}. Differently, when training the generator, we use reconstruction loss, segmentation loss, and generator loss:
\begin{equation}
    \underset{\theta_{G}}{argmin} ~ \mathcal{L}_{recon} + \lambda_{seg}\lambda_{seg} + \lambda_{G}\mathcal{L}_{G}.
    \label{eq:adv_loss}
\end{equation}
We progressively fine-tune the generator following \cite{karras2017progressive}, starting from the $64 \times 64$ render block.

\subsection{Fitting-based Single-Image Face Reconstruction}

Since our model is naturally differentiable, it supports single-image face reconstruction by optimizing (fitting) the style codes $z$s. Given an image $I$, we first use our face encoder to estimate the style code $E(I) = \{ z_{id}, z_{expr}, z_{tex}, z_{light} \}$ as the initial code. The objective is to optimize the code offsets $\Delta z_{g}$ to allow the rendered face image as close as the target face image:
\begin{equation}
    \underset{\{\Delta z_{g} \}}{argmin} ~ \mathcal{L}_{recon}(G(\{ z'_{g} | z'_{g} = z_{g} + \Delta z_{g} \}, p), I),
    \label{eq:fitting_obj}
\end{equation}
where $g \in \{id, expr, tex, light\}$, and $p$ is the estimated camera pose. Note that, the camera poses used during training are estimated by fitting the 3DMM to the images. In fitting-based single-image face reconstruction, we use off-the-shelf face pose estimation network 6DRepNet \cite{hempel20226d} to estimate the yaw, pitch, and roll angles of the face. Because in our framework, we assume the 3D faces are in the canonical pose and the input images are aligned, we can find a transformation to convert the estimated face pose $SO(3)$ to our camera pose $p \in \mathbb{R}^{6}$.

We use gradient descent to optimize the code offsets. The fitting loss is defined as:
\begin{equation}
    \mathcal{L}_{fit} = \mathcal{L}_{recon} + \sum_{g} \| \Delta z_{g} \|^{2}_{2},
    \label{eq:fitting_loss}
\end{equation}
where $\sum_{g} \| \Delta z_{g} \|^{2}_{2}$ is a regularization term to prevent overfitting on a single image.

Note that, the fitting only needs to be done once on a given image, and then we can render the reconstructed face in real-time.

\section{Experiments}
\label{sec:experiments}

\subsection{Dataset}

In this work, we mainly use the Flickr-Faces-HQ (FFHQ) dataset \cite{karras2019style}. The FFHQ dataset contains 70,000 in-the-wild face images with a resolution of 1024$\times$1024. We use an off-the-shelf face-parsing network \footnote{\url{github.com/zllrunning/face-parsing.PyTorch}} to generate the head masks for FFHQ. We use the first 1,000 images as testing data, the next 1,000 images for validation, and the remaining are used for training. The training set has 66,422 images (rather than 68,000) since there are images on which we cannot successfully fit the 3DMM due to issues such as extreme poses. 

We also use the CelebAMask-HQ dataset \cite{CelebAMask-HQ} to test the generalizability of our model. The CelebAMask-HQ dataset contains 30,000 in-the-wild face images of various celebrities with a resolution of $512 \times 512$. It also comes with manually annotated face-parsing masks which we use to derive the binary head masks. We use the same set of 550 randomly selected images for testing as in \cite{galanakis20233dmm}.

\subsection{Implementation Details}

In the multi-resolution photometric loss term, we configure the weighting factors as follows: $\lambda_{64}=0.01$, $\lambda_{128}=0.1$, $\lambda_{256}=0.5$, and $\lambda_{512}=1.0$. For the code loss, we employ weights of $\lambda_{id} = 0.001$, $\lambda_{expr} = 0.1$, $\lambda_{tex} = 0.001$, and $\lambda_{light} = 0.01$. The remaining loss term weights are assigned as: $\lambda_{photo} = 1.0$, $\lambda_{perc} = 0.1$, $\lambda_{seg} = 0.01$, and $\lambda_{G} = 0.1$.

The input images for our encoder adhere to FFHQ's alignment procedure \footnote{\url{https://github.com/NVlabs/ffhq-dataset/blob/master/download_ffhq.py}}, ensuring consistent alignment. Our discriminator network is the same as in StyleGAN2 \cite{karras2020analyzing}. The images are stored and displayed in the 8-bit RGB format. Both input and target images for our network are normalized to fit the intensity range of $[-1.0, 1.0]$. During evaluation, pixel intensities are normalized to the range of $[0, 1.0]$.

In the first training stage, we use an effective batch size of 32 through the gradient accumulation trick (the actual batch size is 8). The number of sampled points along each ray is set to 32 ($N = 32$). We use Adam optimizer with an initial learning rate of $10^{-4}$ to optimize the network parameters. Following each training epoch, the learning rate undergoes a reduction of $2\%$. In the second training stage, we use $N = 64$. The batch size is set to $32$ when the resolution is less than $256 \times 256$, $16$ when the resolution is $256 \times 256$, and $12$ when the resolution is $512 \times 512$. We follow StyleGAN2 to use a learning rate of $10^-4$ for discriminator training, and $2\times 10^{-3}$ for generator training. The training was conducted on one NVIDIA A6000 (48GB VRAM) GPU. The first training stage takes 5 days, and the second training stage takes 6 days.

In single image fitting, we use the Adam optimizer with an initial learning rate of $0.01$ and a decrease rate of $0.99$ per optimization step. We optimize for 200 steps on each image.

\begin{figure*}[h]
    \centering
    \includegraphics[width=\linewidth]{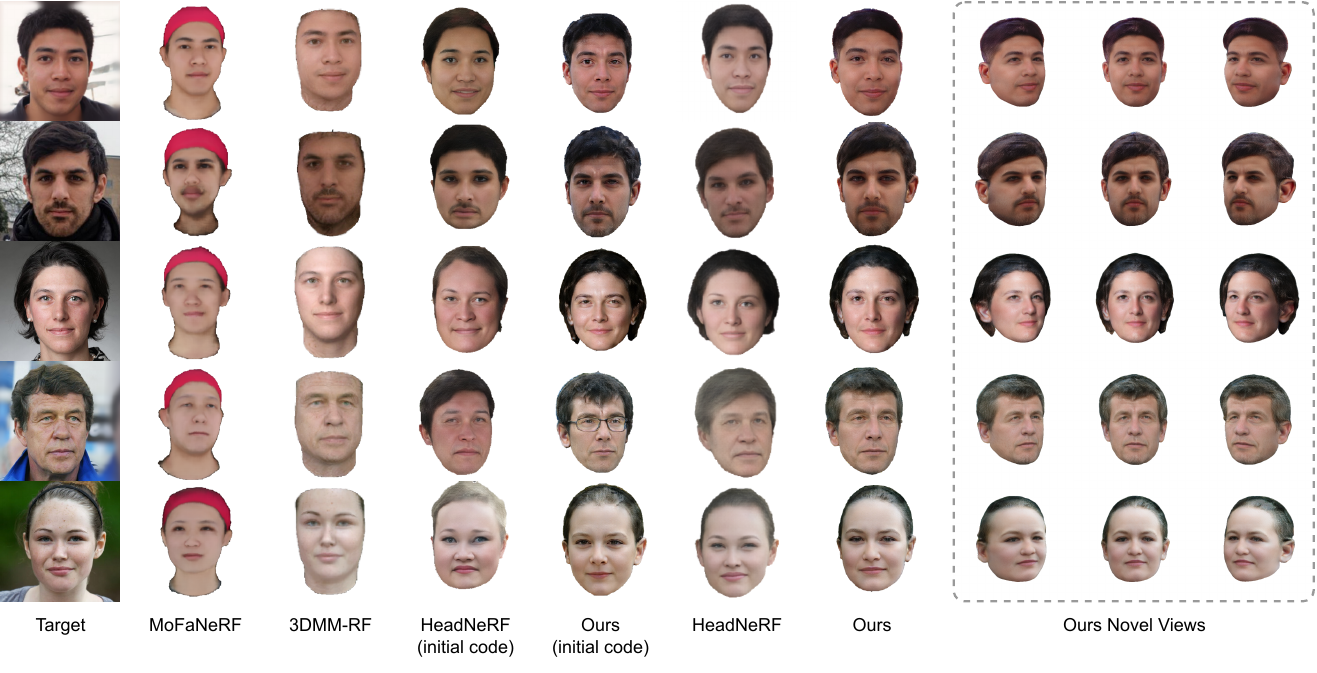}
    \caption{Single image face reconstruction results. The leftmost column shows the target images. We compare with MoFaNeRF \cite{zhuang2022mofanerf}, 3DMM-RF \cite{galanakis20233dmm}, and HeadNeRF \cite{hong2022headnerf}. HeadNeRF (initial code) represents the HeadNeRF's reconstructed face using fitted 3DMM code. Ours (initial code) represents our face reconstruction with our face encoder's estimated style code. The right side shows our reconstructed face rendered in different camera views.}
    \label{fig:recon}
\end{figure*}

\subsection{Single-Image Face Reconstruction}

We evaluate our model in the single-image face reconstruction task to assess its capability in accurately inferring the facial structure and textures from a single input face image. We follow recent works on 3D-aware parametric face models \cite{zhuang2022mofanerf, hong2022headnerf, galanakis20233dmm}, to evaluate our reconstruction quantitatively (through image-level metrics) and qualitatively. We compare StyleMorpheus against MoFaNeRF \cite{zhuang2022mofanerf}, HeadNeRF \cite{hong2022headnerf}, and 3DMM-RF \cite{galanakis20233dmm}. Our approach involves fitting the face encoder's estimated initial code, akin to HeadNeRF's use of fitted 3DMM coefficients as initial code. Thus, we also compare with HeadNeRF, both utilizing initial code. 

\begin{figure}[h]
    \centering
    \includegraphics[width=\linewidth]{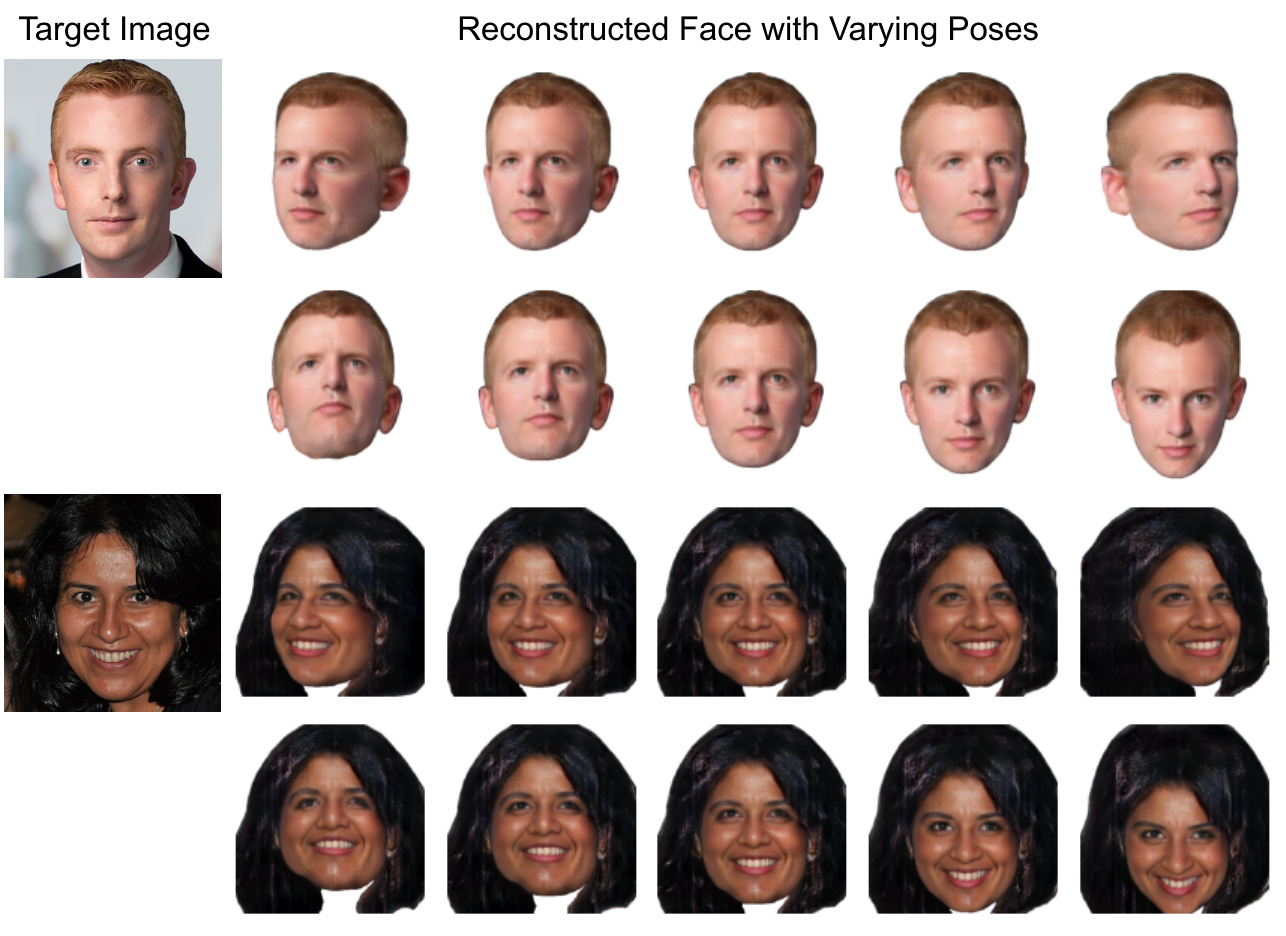}
    \caption{Reconstructed faces viewed from different angles.}
    \label{fig:poses}
\end{figure}

In Fig.~\ref{fig:recon}, we demonstrate the qualitative face reconstruction results on five images collected from the Internet. Confined by training datasets used in MoFaNeRF and 3DMM-RF, both methods can only render parts of the face: MoFaNeRF does not render the hair and 3DMM-RF does not render both hair and ears. Because the 3DMM only models the facial part, its coefficients cannot represent the hair. Thus, HeadNeRF cannot accurately reconstruct the hair using the 3DMM-based initial code. In contrast, our initial code allows our model to render the face close to the target, providing a good starting point for further fine-tuning the code. With the help of adversarial learning, our model can generate more photorealistic details than the other methods. We observe some side effects of adversarial learning in our case. For example, in the fourth row of Fig.~\ref{fig:recon}, our reconstructed face using the initial code has eyeglasses, different from the target image. We interpret this phenomenon caused by the imperfect disentanglement in GANs \cite{chen2022exploring}. We can solve it with code fine-tuning, the eyeglasses disappear, and the final result looks closer to the target image. In Fig.~\ref{fig:poses}, we use more examples to further demonstrate the novel view rendering ability of our model.

We also evaluate the methods quantitatively, through image-level metrics: L1 distance, LPIPS distance \cite{zhang2018unreasonable}, structural similarity index (SSIM), and peak signal-to-noise ratio (PSNR). Table.~\ref{tab:compare-celeb-facial} and ~\ref{tab:compare-celeb-all} show the evaluation on CelebAMask-HQ dataset \cite{CelebAMask-HQ}; Table.~\ref{tab:compare-ffhq-facial} and ~\ref{tab:compare-ffhq-all} show the evaluation on FFHQ dataset \cite{karras2019style}. Our method achieves superior reconstruction performance in terms of image-level similarity.

\begin{table}[ht]
    \definecolor{lg}{rgb}{0.45,0.78,0.65}
    \newcommand{\A}{}
    \newcommand{\B}{}
    \newcommand{\C}{}
    \centering
    \caption{Comparisons in reconstruction performance (Facial Region Only). Evaluated on the CelebAMask-HQ dataset.}
    \begin{tabular}{|l|c|c|c|c|} \hline
        \textbf{Method} & \textbf{L1} $\downarrow$ & \textbf{LPIPS} $\downarrow$ & \textbf{SSIM} $\uparrow$ & \textbf{PSNR} $\uparrow$ \\\hline

        MoFaNeRF  & 0.273 & 0.442 & 0.910 & 14.713 \\\hline
        
        3DMM-RF*  & 0.216 & - & \textbf{0.956}\A & - \\\hline

        HeadNeRF (initial code) & 0.086 & 0.133 & 0.928 & 23.441 \\\hline 

        HeadNeRF  & 0.072\C & 0.113\C & 0.940\C & 24.937\B \\\hline 

        Ours (initial code)     & 0.068\B & 0.112\B & 0.927 & 24.619\C \\\hline 

        Ours                    & \textbf{0.048}\A & \textbf{0.094}\A & 0.951\B & \textbf{27.586}\A \\\hline 

        \multicolumn{5}{l}{* indicates results reported from the original paper (not open-source).}\\ 
        \multicolumn{5}{l}{- indicates the value is missing or not comparable.}\\
    \end{tabular}
    \label{tab:compare-celeb-facial}

    \vspace{0.5cm}

    \centering
    \caption{Comparisons in reconstruction performance (Including Hair). Evaluated on the CelebAMask-HQ dataset.}
    \begin{tabular}{|l|c|c|c|c|} \hline
        \textbf{Method} & \textbf{L1} $\downarrow$ & \textbf{LPIPS} $\downarrow$ & \textbf{SSIM} $\uparrow$ & \textbf{PSNR} $\uparrow$ \\\hline

        HeadNeRF (initial code) & 0.366 & 0.472 & 0.675 & 14.926 \\\hline 

        HeadNeRF & 0.354\C & 0.431\C & 0.693\C & 15.429\C \\\hline 

        Ours (initial code)     & 0.208\B & 0.355\B & 0.702\B & 18.619\B \\\hline 

        Ours                    & \textbf{0.161}\A & \textbf{0.313}\A & \textbf{0.742}\A & \textbf{21.041}\A \\\hline 


    \end{tabular}
    \label{tab:compare-celeb-all}
\end{table}

\begin{table}[ht]
    \definecolor{lg}{rgb}{0.45,0.78,0.65}
    \newcommand{\A}{}
    \newcommand{\B}{}
    \newcommand{\C}{}
    \centering
    \caption{Comparisons in reconstruction performance (Facial Region Only). Evaluated on FFHQ dataset.}
    \begin{tabular}{|l|c|c|c|c|} \hline
        \textbf{Method} & \textbf{L1} $\downarrow$ & \textbf{LPIPS} $\downarrow$ & \textbf{SSIM} $\uparrow$ & \textbf{PSNR} $\uparrow$ \\\hline

        MoFaNeRF  & 0.251 & 0.416 & 0.720 & 15.084 \\\hline
        
        HeadNeRF (initial code) & 0.106 & 0.171 & 0.902\C & 22.211 \\\hline 

        HeadNeRF  & 0.087\C & 0.151\B & 0.920\B & 23.858\B \\\hline 

        Ours (initial code)     & 0.085\B & 0.152\C & 0.900 & 23.533\C \\\hline 

        Ours                    & \textbf{0.063}\A & \textbf{0.124}\A & \textbf{0.922}\A & \textbf{26.038}\A \\\hline 

        
    \end{tabular}
    \label{tab:compare-ffhq-facial}

    \vspace{0.5cm}

    \centering
    \caption{Comparisons in reconstruction performance (Including Hair). Evaluated on FFHQ dataset.}
    \begin{tabular}{|l|c|c|c|c|} \hline
        \textbf{Method} & \textbf{L1} $\downarrow$ & \textbf{LPIPS} $\downarrow$ & \textbf{SSIM} $\uparrow$ & \textbf{PSNR} $\uparrow$ \\\hline

        HeadNeRF (initial code) & 0.241 & 0.355 & 0.774 & 17.499 \\\hline 

        HeadNeRF  & 0.211\C & 0.310\C & 0.798\B & 18.794\C \\\hline 

        Ours (initial code)     & 0.160\B & 0.281\B & 0.781\C & 20.254\B \\\hline 

        Ours                    & \textbf{0.126}\A & \textbf{0.257}\A & \textbf{0.814}\A & \textbf{22.269}\A \\\hline 


    \end{tabular}
    \label{tab:compare-ffhq-all}
\end{table}

\begin{figure*}[ht]
    \centering
    \includegraphics[width=\linewidth]{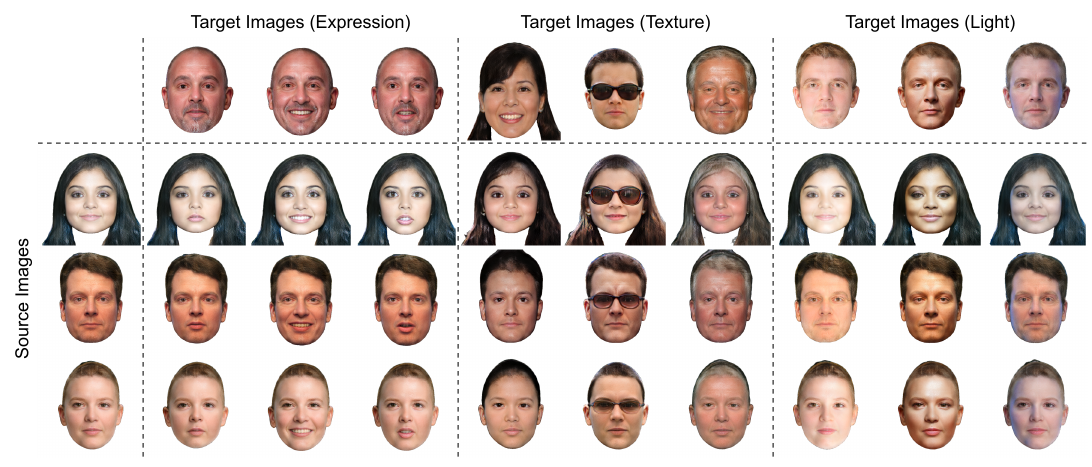}
    \caption{Disentangled style mixing of latent space. The leftmost column shows source images and the first row shows the target images. We divide the target images into three groups: expression, texture, and light. The target image provides the corresponding semantic group of code, and the source image provides the rest semantic groups of code.}
    \label{fig:disentangle}
\end{figure*}

\subsection{Applications}

We implemented multiple applications to further showcase the potential of StyleMorpheus. 

\subsubsection{Style-Mixing}

Inspired by style-based image generators \cite{karras2019style, gu2021stylenerf}, we employ style mixing to showcase our model's disentangled control capability. In our case, we mix the $z$ style codes. We define source image $I_{src}$ and target image $I'_{tgt}$. Their fitted style codes are $z_{src} = \{ z_{id}, z_{expr}, z_{tex}, z_{light} \}$ for source image $I_{src}$ and $z'_{tgt} = \{ z'_{id}, z'_{expr}, z'_{tex}, z'_{light} \}$ for target image $I'_{tgt}$. We replace one segment of the source code with its corresponding target code, to get the mixed code $z^{*}$. For example, if we want to use the target to provide the facial expression (expr.), then we replace $z_{expr}$ with $z'_{expr}$: $z^{*} = \{ z_{id}, z'_{expr}, z_{tex}, z_{light} \}$. Fig.~\ref{fig:disentangle} shows the style mixing results.




\subsubsection{Facial Part Color Editing}

\begin{figure}[ht]
    \centering
    \includegraphics[width=\linewidth]{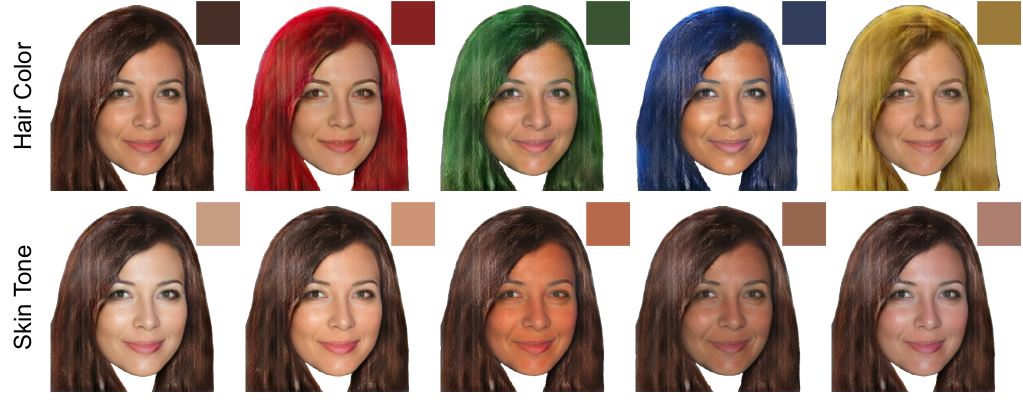}
    \caption{Part color editing examples. In the first row, we edit the hair color; in the second row, we edit the skin tone. The average color of the corresponding part is shown in the top right patch.}
    \label{fig:color_editing}
\end{figure}

\begin{figure*}[ht]
    \centering
    \includegraphics[width=\linewidth]{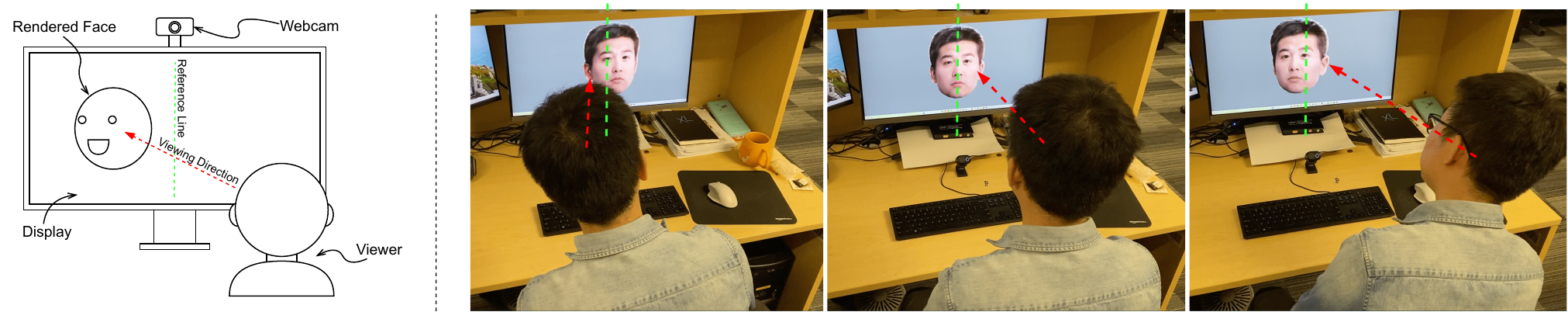}
    \caption{Our virtual reality demonstration. The left side shows the setup of our demonstration. It includes a high-resolution display, a calibrated webcam, and a viewer. The viewer sits in front of the display. When the viewer's head moves, we render the face using the corresponding camera pose to create the illusion of a 3D effect. The right side shows the rendered faces viewed from different directions. The \textcolor{green}{green} dashed line indicates the horizontal center of the screen. The \textcolor{red}{red} dashed line with an arrow indicates the viewer's viewing direction.}
    \label{fig:holo}
\end{figure*}

We propose a novel approach for editing the color of facial parts such as hair and skin, using StyleMorpheus. Starting with the face code $z = \{ z_{id}, z_{expr}, z_{tex}, z_{light} \} $ and a camera pose $p$, we render the facial image $I = G(z, p)$. To edit the part color, we first run the face parsing model on $I$, to get the corresponding part's binary mask $\mathbf{M}_{part}$. Then, we define the texture code offsets $\Delta z_{tex}$ (initialized as zeros), to form the edited face code $z' = \{ z_{id}, z_{expr}, z_{tex} + \Delta z_{tex}, z_{light} \}$. We render the edited image as $I' = G(z', p) $. We define an image $I^{*}$ that has the same shape as $I$ and $I'$ to have pure color (all the pixels have the same intensities) as the target part color. Our objective is then formulated as:
\begin{equation}
    \underset{\Delta z_{tex}}{argmin} ~ \mathcal{L}_{color} + \mathcal{L}_{rest} + \| \Delta z_{tex} \|^2_2, 
    \label{eq:color_editing}
\end{equation}
where
\begin{equation}
    \mathcal{L}_{color} = \frac{\| I' \odot \mathbf{M}_{part} - I^* \odot \mathbf{M}_{part} \|^2_2}{ \| \mathbf{M}_{part} \|_{1} } , 
\end{equation}
\begin{equation}
    \mathcal{L}_{rest} = \frac{ \| I' \odot (1 - \mathbf{M}_{part}) - I \odot (1 - \mathbf{M}_{part}) \|^2_2 }{ \| 1 - \mathbf{M}_{part} \|_{1} }, 
\end{equation}
and $\| \Delta z_{tex} \|^2_2$ is a regularization term. The purpose of $\mathcal{L}_{color}$ is to allow the average part color to approach the target part color. Both $\mathcal{L}_{rest}$ and $\| \Delta z_{tex} \|^2_2$ help to regulate the code offset from being too large and to ensure the edited face still resembles the original face. Fig.~\ref{fig:color_editing} shows our part color editing results. Our face part color editing allows the user to further customize a reconstructed face.

\subsubsection{Real-Time Rendering and Virtual Reality Demonstration}

Real-time performance is crucial in virtual reality applications to provide seamless interactions and user immersion, enhancing the overall experience and preventing discomfort. Olano et al. suggest that the frames-per-second (FPS) should be not less than 15 to be considered as real-time rendering \cite{olano2004real}. We test several 3D-aware face models on our devices to benchmark their ability in real-time rendering. For a fair comparison, we set the rendered image size to $512 \times 512$, and the number of points samples per ray to be $64$. We benchmark on two devices. The first device is a personal computer with an Intel i7-11700F CPU, 32GB RAM, and a mid-range NVIDIA RTX 3070 (8GB VRAM) GPU. The second device is a server computer with an AMD Threadripper pro 5995wx CPU, 512GB RAM, and an NVIDIA RTX A6000 (48GB VRAM) GPU. The benchmark results are shown in Table~\ref{tab:fps}. As we can see, our method has achieved satisfactory real-time performance even on a personal computer. Methods include StyleNeRF \cite{gu2021stylenerf}, MoFaNeRF \cite{zhuang2022mofanerf}, and Next3D \cite{sun2023next3d} fail to pass our real-time benchmark on the mid-range GPU. Because MoFaNeRF is basically an end-to-end NeRF, it is computationally expensive especially when rendering high-resolution images.

\begin{table}[ht]

    \small
    \centering
    \caption{Frames-per-second $\uparrow$ of different 3D-aware face models on different devices.}
    \begin{tabular}{l|c|c} \hline
         \textbf{Method} & \makecell{\textbf{RTX 3070}\\(8GB VRAM)} & \makecell{\textbf{RTX A6000}\\(48GB VRAM)} \\\hline
        StyleNeRF \cite{gu2021stylenerf} & \textcolor{red}{5}   & \textcolor{red}{8} \\
        MoFaNeRF \cite{zhuang2022mofanerf} & \textcolor{red}{Out of memory} & \textcolor{red}{2} \\
        Next3D  \cite{sun2023next3d}  & \textcolor{red}{11}  & 19 \\
        HeadNeRF \cite{hong2022headnerf} & 24  & \textbf{56} \\
        Ours      & \textbf{26} & 53 \\\hline
        \multicolumn{3}{l}{The \textcolor{red}{red} color indicates that real-time has not been reached.} 
    \end{tabular}
    \label{tab:fps}
\end{table}

We developed a simple virtual reality application for demonstrating the proposed StyleMorpheus in a real-time 3D-aware image synthesis setting. As illustrated in the left side of Fig.~\ref{fig:holo}, our setup comprises a high-resolution computer display, a webcam, and a person as the viewer. We use the webcam to capture the viewer's face and estimate its pose (including its relative location to the center of the screen). This estimated pose is then used to determine the camera's perspective for rendering the face. As the viewer's pose changes, the rendered face pose and location adjust accordingly. This creates the illusion of peering through a window to perceive a genuine 3D face in physical space. The right side of Fig.~\ref{fig:holo} depicts some examples of the rendered face on the display. This demonstration shows the potential of our StyleMorpheus in immersive virtual reality applications, such as telepresence.

\subsection{Ablation Study}

We conduct the ablation study to gain insights into components of our design's impact on the model's overall performance. We organize our ablation study into two experiments. The first experiment studies the impact of loss terms in our training loss function. The second experiment studies the importance of using a face encoder to learn the style code space.

\begin{figure}[ht]
    \centering
    \includegraphics[width=\linewidth]{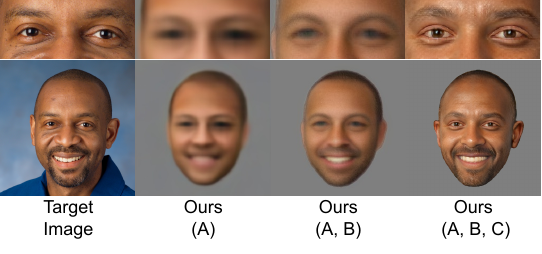}
    \caption{Ablation study on the loss function. \textbf{A}: $\mathcal{L}_{photo}$; \textbf{B}: $\mathcal{L}_{perc}$; \textbf{C}: adversarial loss.}
    \label{fig:ablation_loss}
\end{figure}

\begin{figure}[ht]
    \centering
    \includegraphics[width=\linewidth]{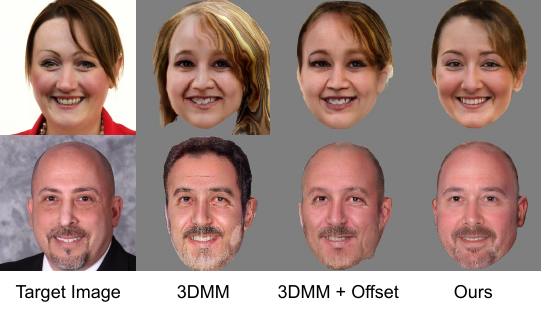}
    \caption{Ablation study on the input code.}
    \label{fig:ablation_code}
\end{figure}

In the first experiment, we study the contribution of our loss terms in rendered image quality. In the reconstruction loss, we apply the photometric loss term $\mathcal{L}_{photo}$ and perceptual loss term $\mathcal{L}_{perc}$. The reconstruction loss is used in both of our training stages. In the second stage, we used adversarial learning. The adversarial learning alternately optimizes $G$ and $D$, with corresponding loss terms. For convenience, we define adversarial learning loss as both $\mathcal{L}_{G}$ and $\mathcal{L}_{D}$. Fig.~\ref{fig:ablation_loss} shows the example reconstruction results using our model trained with different loss combinations. Since $\mathcal{L}_{photo}$ is the most used loss term in face reconstruction works, we test only training our model on $\mathcal{L}_{photo}$, see Fig.~\ref{fig:ablation_loss} (A). It can be observed that when we only use $\mathcal{L}_{photo}$, the reconstructed result appears very blurry, and the identity does not match the target image well. When we use both $\mathcal{L}_{photo}$ and $\mathcal{L}_{perc}$ in training our model, see Fig.~\ref{fig:ablation_code} (A, B), the reconstructed face looks closer to the target image. However, the reconstruction is still blurry. We believe the main reasons are the estimated camera poses we use are not accurate enough and the complexity of in-the-wild images. When we add the adversarial learning loss, the reconstruction quality improved significantly, see Fig.~\ref{fig:ablation_code} (A, B, C). We also test the three models on the CelebAMask-HQ dataset, results are shown in Table~\ref{tab:ablation_loss}. Quantitative results are consistent with the visual quality, suggesting that all the above-mentioned loss terms are important in our model training.  

\begin{table}[ht]
    \definecolor{lg}{rgb}{0.45,0.78,0.65}
    \newcommand{\A}{\cellcolor{lg!100}}
    \newcommand{\B}{\cellcolor{lg!50}}
    \newcommand{\C}{\cellcolor{lg!20}}
    
    \centering
    \caption{Ablation study on the loss function.}
    \begin{tabular}{|c|c|c|c|c|c|c|} \hline
         \multicolumn{3}{|c|}{\textbf{Loss Terms}} & \multicolumn{4}{c|}{\textbf{Evaluation Metrics}} \\\hline
         \textbf{A} & \textbf{B} & \textbf{C} & \textbf{L1} $\downarrow$ & \textbf{LPIPS} $\downarrow$ & \textbf{SSIM} $\uparrow$ & \textbf{PSNR} $\uparrow$
         \\\hline
         \cmark & \xmark & \xmark  & 0.322 & 0.494 & 0.731 & 15.341
         \\\hline
         \cmark & \cmark & \xmark & 0.281 & 0.409 & 0.681 & 15.824
         \\\hline
         \cmark & \cmark & \cmark & \textbf{0.126} & \textbf{0.257} & \textbf{0.814} & \textbf{22.269}
         \\\hline
         \multicolumn{7}{l}{\textbf{A}: $\mathcal{L}_{photo}$; \textbf{B}: $\mathcal{L}_{perc}$; \textbf{C}: adversarial loss.}
    \end{tabular}
    \label{tab:ablation_loss}
\end{table}

\begin{figure}[ht]
    \centering
    \includegraphics[width=.7\linewidth]{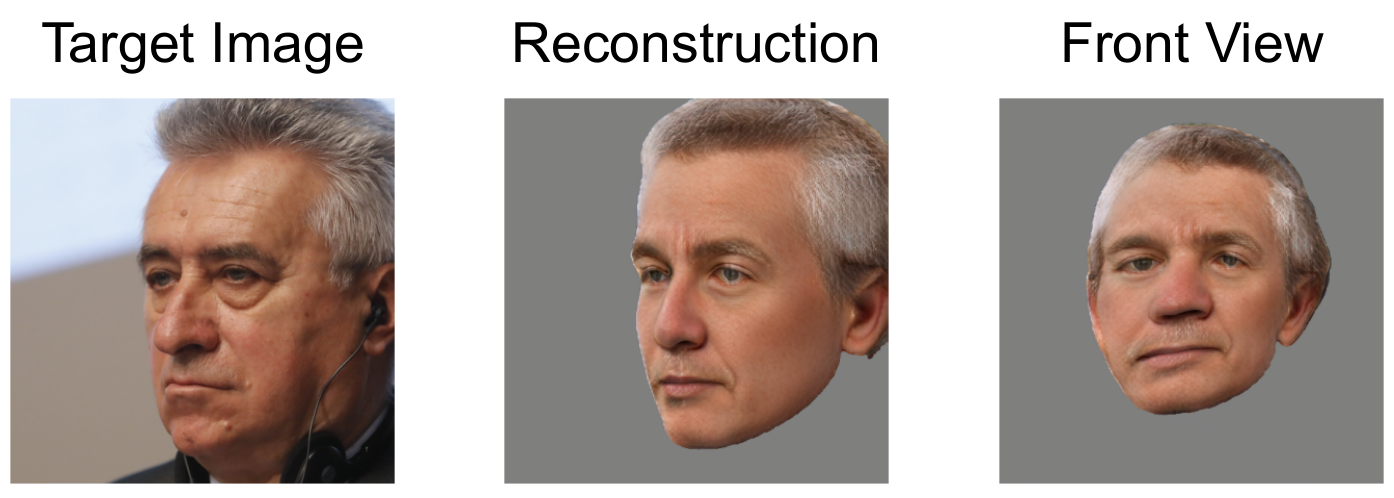}
    \caption{An example of the failure case. When the estimated camera pose is not accurate enough, the reconstructed face gets distorted in other viewing directions.}
    \label{fig:limitation}
\end{figure}

In our second experiment, we examine the use of fixed input codes for our generator, indicating that the face encoder is not involved. The outcomes of this experiment are illustrated in Fig.~\ref{fig:ablation_code}. We initiate by employing the fitted 3DMM code as input. It is evident that the reconstructed faces exhibit inaccurate hairstyles due to the restricted representation within the 3DMM code space. Subsequently, we employ HeadNeRF's improved 3DMM code (3DMM + offset) as input. Given that HeadNeRF fine-tunes the 3DMM code by incorporating offsets to encompass more information, it can approximately capture the hairstyles. Nevertheless, using fixed input codes results in notable artifacts in the generated images. We also test the reconstruction on the CelebAMask-HQ dataset, results are shown in Table~\ref{tab:ablation_code}.

\begin{table}[ht]
    \centering
    \caption{Ablation study on the input style code.}
    \begin{tabular}{|l|c|c|c|c|} \hline
         \textbf{Input Code} & \textbf{L1} $\downarrow$ & \textbf{LPIPS} $\downarrow$ & \textbf{SSIM} $\uparrow$ & \textbf{PSNR} $\uparrow$
         \\\hline
         3DMM & 0.407 & 0.466 & 0.612 & 13.074   
         \\\hline
         3DMM + Offset & 0.282 & 0.406 & 0.675 & 15.771
         \\\hline
         Ours & \textbf{0.126} & \textbf{0.257} & \textbf{0.814} & \textbf{22.269}
         \\\hline

    \end{tabular}
    \label{tab:ablation_code}
\end{table}

\subsection{Limitation and Future Work}


Through our experiment, we identified some limitations with our current model. In rare cases of single-image face reconstruction, when the face pose is large and the estimated camera pose is not accurate enough, the reconstruction tends to be inaccurate. This inaccurate reconstruction also leads to a distorted face if we render it in other views (see Fig.~\ref{fig:limitation}). In our future work, we plan to study the simultaneous optimization of camera pose and reconstruction. Another limitation is in the exaggerated facial expressions. Since the in-the-wild images we use do not cover a lot of exaggerated facial expressions, our model usually fails in reconstructing face images with exaggerated facial expressions. To tackle this challenge, we plan to leverage in-the-wild video face datasets, such as VFHQ \cite{xie2022vfhq}, to improve the representation ability of facial expressions. We will also explore fine-tuning our model for more personalized face representation.

\section{Conclusion}
\label{sec:conclusion}

This paper introduces StyleMorpheus, the first style-based neural 3DMM trained on in-the-wild data that enables the reconstruction of high-fidelity faces with editable attributes from single-view images. The approach involves a face auto-encoder for learning a disentangled facial representation space, allowing for independent manipulation of attributes. We design a style-based 3D-aware decoder conditioned on encoded face representation to render facial images in varying views. Through adversarial learning, StyleMorpheus excels in capturing intricate facial details, resulting in impressive photorealistic reconstructions solely trained on in-the-wild data. StyleMorpheus outperforms existing 3D-aware parametric face models in terms of reconstruction quality.  Furthermore, we have implemented several practical applications to showcase the real-world utility of StyleMorpheus.



\ifCLASSOPTIONcaptionsoff
  \newpage
\fi



\bibliographystyle{IEEEtran}
\bibliography{references}
%

%

\begin{IEEEbiography}[{\includegraphics[width=1in,height=1.25in,clip,keepaspectratio]{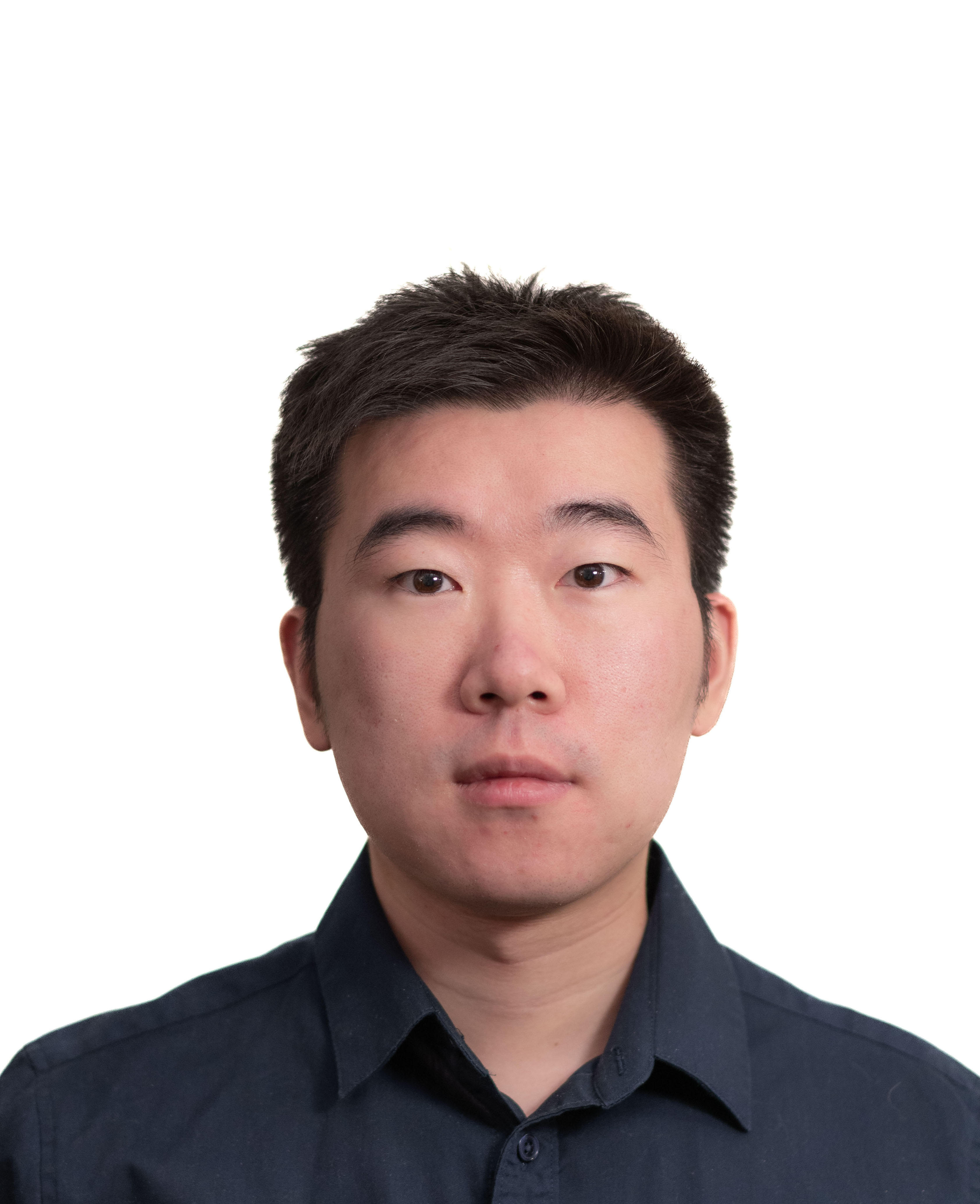}}]{Peizhi Yan} is a Ph.D candidate at the University of British Columbia, BC, Canada, as of 2022. He received M.Sc. in Computer Science (2020) from Lakehead University, Thunder Bay, ON, Canada; B.Sc. in Computer Science (2018) from Algoma University, Sault Ste. Marie, ON, Canada; and B.Sc. in Computer Science (2019) from University of Jinan, Jinan, Shandong, China. His research interests are 3D computer vision and deep learning. Peizhi was the recipient of the Canadian Governor General’s Academic Gold Medal and the recipient of the Toronto Vector Institute's first round of Vector Scholarship in Artificial Intelligence. He also serves as a reviewer of many journals.
\end{IEEEbiography}

\begin{IEEEbiography}[{\includegraphics[width=1in,height=1.25in,clip,keepaspectratio]{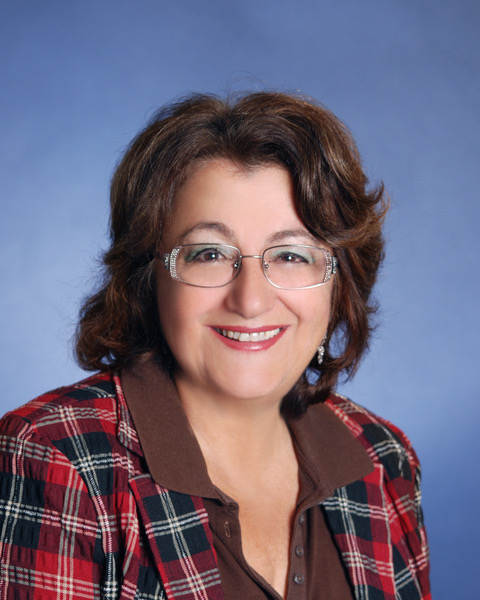}}]{Rabab Kreidieh Ward} has over 40 years of post-doctoral experience in education, research, development and leadership. Her research interests are in broad areas of signal and image processing and their applications. She has published around 600 refereed journal and conference papers and holds ten patents. Some of her work has been licensed to US and Canadian industries. She is a member of the (USA) National Academy of Engineering, Fellow of the Royal Society of Canada, the IEEE, the Canadian Academy of Engineers, and the Engineering Institute of Canada. Amongst her several awards are the  IEEE Fourier Award for Signal Processing,  the  Norbert Wiener Society Award of the IEEE  Signal Processing Society, and UBC Killam Research Prize. 
\end{IEEEbiography}


\begin{IEEEbiography}[{\includegraphics[width=1in,height=1.25in,clip,keepaspectratio]{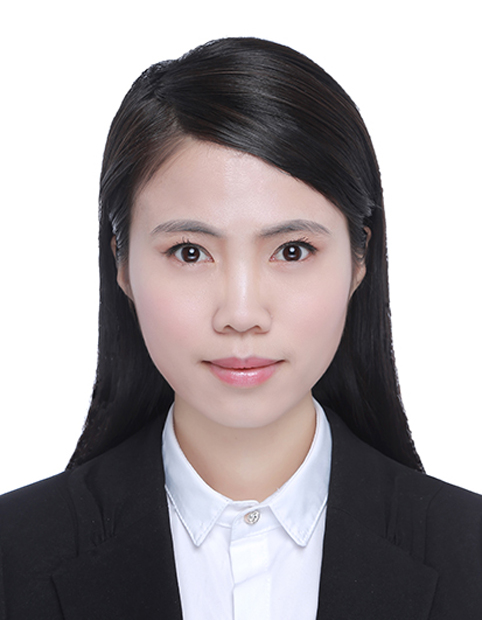}}]{Dan Wang} received the B.S. and M.Sc. degrees from Beihang University, Beijing, China. She is currently pursuing the Ph.D. degree at the University of British Columbia, Vancouver, BC, Canada. Her research fields include explainable artificial intelligence and 3D neural rendering.
\end{IEEEbiography}

\begin{IEEEbiography}[{\includegraphics[width=1in,height=1.25in,clip,keepaspectratio]{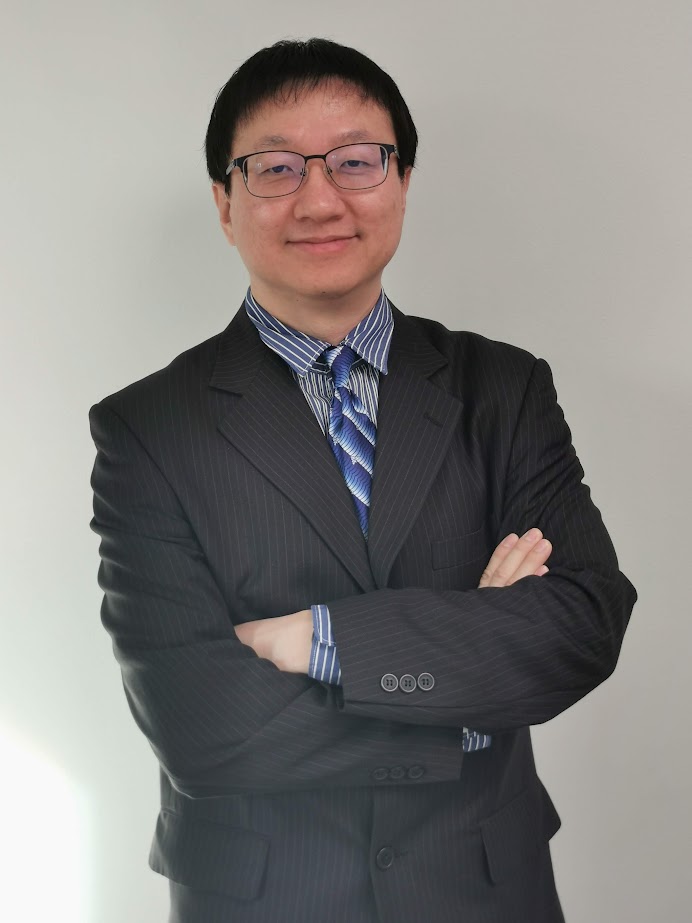}}]{Qiang Tang} is currently a Principal Engineer at Huawei Technologies Canada Co. Ltd. He received the B.S. and M.S. degrees in electrical engineering from Tianjin University, in 2001 and 2004, respectively. He received his Ph.D. degree (2010) in electrical and computer engineering at the University of British Columbia. Before joining Huawei, he was a senior research engineer at Point Grey Research Inc. He has been working on projects across a broad range of fields, which include multi-modal large language models, autonomous driving, neural rendering, 3D reconstruction, 3D content generation, image sensor pipeline, image/video segmentation, and image/video compression.  
\end{IEEEbiography}

\begin{IEEEbiography}[{\includegraphics[width=1in,height=1.25in,clip,keepaspectratio]{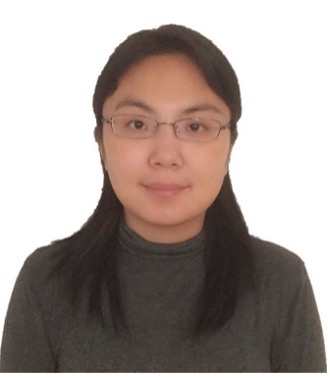}}]{Shan Du} (S’05-M’09-SM’16) received the Ph.D. degree in Electrical and Computer Engineering from the University of British Columbia, Vancouver, BC, Canada in 2009. She is currently an assistant professor with the Department of Computer Science, Mathematics, Physics \& Statistics, University of British Columbia (Okanagan Campus). Before joining UBC, she worked as an assistant professor with the Department of Computer Science, Lakehead University, Canada, and as a Research Scientist/Software Engineer with IntelliView Technologies Inc., Canada. Shan has more than 15 years of research and development experience in image/video processing, image/video analysis, computer vision, pattern recognition, and machine learning. Shan was the recipient of many awards and grants, including NSERC-IRDF, NSERC-CGS D, AITF Industry r\&D Associates Grant, ICASSP Best Paper Award, NSERC DG, CFI JELF, etc. Shan is a senior member of IEEE, IEEE Signal Processing Society, and IEEE Circuits and Systems Society. She is serving as an Associate Editor of IEEE Trans. on Circuits and Systems for Video Technology and IEEE Canadian Journal of Electrical and Computer Engineering, Area Chair of ICIP 2023, and served as Area/Session Chair of ICIP 2022, 2021, and 2019, TPC member and reviewer for many international journals and conferences.
\end{IEEEbiography}




\vfill

\end{document}